\def\year{2018}\relax
\documentclass[letterpaper]{article} 
\usepackage{aaai18}  
\usepackage{times}  
\usepackage{helvet}  
\usepackage{courier}  
\usepackage{url}  
\usepackage{graphicx}  
\usepackage{wrapfig}
\usepackage{amssymb}
\usepackage{amsmath}
\usepackage{subfigure}
\usepackage{multirow}
\usepackage{float}
\usepackage{algorithm}
\usepackage{algorithmic}

\def \f {\mathbf{f}}
\def \F {\mathbf{F}}

\def \x {\mathbf{x}}

\def \h {\mathbf{h}}

\def \D {\mathcal{D}}

\def \I {\mathbb{I}}

\def \bq {\begin{eqnarray}}
\def \eq {\end{eqnarray}}
\def \bqs {\begin{eqnarray*}}
	\def \eqs {\end{eqnarray*}}

\begin{document}
%
\title{Online Deep Learning: Learning Deep Neural Networks on the Fly}
\author{Doyen Sahoo, Quang Pham, Jing Lu, Steven C.H. Hoi\\
School of Information Systems, Singapore Management Univeristy\\
\{doyens,hqpham,jing.lu.2014,chhoi\}@smu.edu.sg
}
\maketitle
\begin{abstract}
Deep Neural Networks (DNNs) are typically trained by backpropagation in a batch learning setting, which requires the entire training data to be made available prior to the learning task. This is not scalable for many real-world scenarios where new data arrives sequentially in a stream form. We aim to address an open challenge of ``Online Deep Learning" (ODL) for learning DNNs on the fly in an online setting. Unlike traditional online learning that often optimizes some convex objective function with respect to a shallow model (e.g., a linear/kernel-based hypothesis), ODL is significantly more challenging since the optimization of the DNN objective function is non-convex, and regular backpropagation does not work well in practice, especially for online learning settings. In this paper, we present a new online deep learning framework that attempts to tackle the challenges by learning DNN models of adaptive depth from a sequence of training data in an online learning setting. In particular, we propose a novel Hedge Backpropagation (HBP) method for online updating the parameters of DNN effectively, and validate the efficacy of our method on large-scale data sets, including both stationary and concept drifting scenarios. 
\end{abstract}

\section{Introduction}
Recent years have witnessed tremendous success of deep learning techniques in a wide range of applications \cite{lecun2015deep,bengio2013representation,bengio2015deep,krizhevsky2012imagenet,he2015deep}.
Learning Deep Neural Networks (DNN) faces many challenges, including (but not limited to) vanishing gradient, diminishing feature reuse \cite{srivastava2015training}, saddle points (and local minima) \cite{choromanska2015loss,dauphin2014identifying}, immense number of parameters to be tuned, internal covariate shift during training \cite{ioffe2015batch}, difficulties in choosing a good regularizer, choosing hyperparameters, etc. Despite many promising advances \cite{nair2010rectified,ioffe2015batch,he2015deep,srivastava2015training}, etc., which are designed to address specific problems for optimizing deep neural networks, most of these existing approaches assume that the DNN models are trained in a batch learning setting which requires the entire training data set to be made available prior to the learning task. This is not possible for many real world tasks where data arrives sequentially in a stream, and may be too large to be stored in memory. Moreover, the data may exhibit concept drift \cite{gamajo2014survey}. Thus, a more desired option is to learn the models in an online setting.

Unlike batch learning, online learning \cite{Zinkevich2003,Cesa-Bianchi2006} represents a class of learning algorithms that learn to optimize predictive models over a stream of data instances sequentially. The on-the-fly learning makes online learning highly scalable and memory efficient. However, most existing online learning algorithms are designed to learn shallow models (e.g., linear or kernel methods \cite{Crammer2006,Kivinen2004,Hoi2013}) with online convex optimization, which cannot learn complex nonlinear functions in complicated application scenarios.

In this work, we attempt to bridge the gap between online learning and deep learning by addressing the open problem of ``Online Deep Learning" (ODL) --- how to learn Deep Neural Networks (DNNs) from data streams in an online setting. A possible way to do ODL is to put the process of training DNNs online by directly applying a standard Backpropagation training on only a single instance at each online round. Such an approach is simple but falls short due to some critical limitations in practice. One key challenge is how to choose a proper model capacity (e.g., depth of the network) before starting to learn the DNN online. If the model is too complex (e.g., very deep networks), the learning process will converge too slowly (vanishing gradient and diminishing feature reuse), thus losing the desired property of online learning. On the other extreme, if the model is too simple, the learning capacity will be too restricted, and without the power of depth, it would be difficult to learn complex patterns. In batch learning literature, a common way to address this issue is to do model selection on validation data. Unfortunately, it is not realistic to have validation data in online settings, and is thus infeasible to apply traditional model selection in online learning scenarios. In this work, we present a novel framework for online deep learning, which is able to learn DNN models from data streams sequentially, and more importantly, is able to adapt its model capacity from simple to complex over time, nicely combining the merits of both online learning and deep learning.

We aim to devise an online learning algorithm that is able to start with a shallow network that enjoys fast convergence; then gradually switch to a deeper model (meanwhile sharing certain knowledge with the shallow ones) automatically when more data has been received to learn more complex hypotheses, and effectively improve online predictive performance by adapting the capacity of DNNs. To achieve this, we need to address questions such as: \textit{when} to change the capacity of network? \textit{how} to change the capacity of the network? and how to do both in an \textit{online} setting? We design an elegant solution to do all this in a unified framework in a data-driven manner. We first amend the existing DNN architecture by attaching every hidden layer representation to an output classifier. Then, instead of using a standard Backpropagation, we propose a novel \emph{Hedge Backpropagation} method, which evaluates the online performance of every output classifier at each online round, and extends the Backpropagation algorithm to train the DNNs online by exploiting the classifiers of different depths with the Hedge algorithm \cite{Freund1997}. This allows us to train DNNs of adaptive capacity meanwhile enabling knowledge sharing between shallow and deep networks.

\section{Related Work}

\subsection{Online Learning}
Online Learning represents a family of scalable and efficient algorithms that learn to update models from data streams sequentially \cite{Cesa-Bianchi2006,Shalev-Shwartz2007,hoi2014libol,wu2017sol}. Many techniques are based on maximum-margin classification, from Perceptron \cite{Rosenblatt1958} to Online Gradient Descent \cite{Zinkevich2003}, Passive Aggressive (PA) algorithms \cite{Crammer2006}, Confidence-Weighted (CW) Algorithms, \cite{dredze2008confidence} etc. These are primarily designed to learn linear models. Online Learning with kernels \cite{Kivinen2004} offered a solution for online learning with nonlinear models. These methods received substantial interest from the community, and models of higher capacity such as Online Multiple Kernel Learning were developed \cite{Jin2010,Hoi2013,Sahoo2014,lu2015budget,sahoo2016cost}. While these models learn nonlinearity, they are still shallow. Moreover, deciding the number and type of kernels is non-trivial; and these methods are not explicitly designed to \emph{learn} a feature representation.

Online Learning can be directly applied to DNNs ("online backpropagation") but they suffer from convergence issues, such as vanishing gradient and diminishing feature reuse. Moreover, the optimal depth to be used for the network is usually unknown, and cannot be validated easily in the online setting. 
Further, networks of different depth would be suitable for different number of instances to be processed, e.g., for small number of instances - a quick convergence would be of high priority, and thus shallow networks would be preferred, whereas, for a large number of instances, the long run performance would be enhanced by using a deeper network. This makes model architecture selection very challenging. There have been attempts at making deep learning compatible with online learning \cite{zhou2012online,lee2016dual} and \cite{lee2017lifelong}. However, they operate via a sliding window approach with a (mini)batch training stage, making them unsuitable for a streaming data setting.

\subsection{Deep Learning}

Due to the difficulty in training deep networks, there has been a large body of emerging works adopting the principle of "shallow to deep". This is similar to the principle we adopt in our work. This approach exploits the intuition that shallow models converge faster than deeper models, and this idea has been executed in several ways. Some approaches do this explicitly by Growing of Networks via the function preservation principle \cite{chen2015net2net,wei2016network}, where the (student) network of higher capacity is guaranteed to be at least as good as the shallower (teacher) network. Other approaches perform this more implicitly by modifying the network architecture and objective functions to enable the network to allow the input to flow through the network, and slowly adapt to deep representation learning, e.g., Highway Nets\cite{srivastava2015training}, Residual Nets\cite{he2015deep}, Stochastic Depth Networks \cite{huang2016deep} and Fractal Nets \cite{larsson2016fractalnet}.

However, they are all designed to optimize the loss function based on the output obtained from the deepest layer. Despite the improved batch convergence, they cannot yield good online performances (particularly for the instances observed in early part of the stream), as the inference made by the deepest layer requires substantial time for convergence. For the online setting, such existing deep learning techniques could be trivially beaten by a very shallow network. Deeply Supervised Nets \cite{lee2015deeply}, shares a similar architecture as ours, which uses companion objectives at every layer to address vanishing gradient and to learn more discriminative features at shallow layers. However, the weights of companions are set heuristically, where the primary goal is to optimize the classification performance based on features learnt by the deepest hidden layer, making it suitable only for batch settings, which suffers from the same drawbacks as others.

Recent years have also witnessed efforts in learning the architecture of the neural networks \cite{srinivas2015learning,alvarez2016learning}, which incorporate architecture hyperparameters into the optimization objective. Starting from an overcomplete network, they use regularizers that help in eliminating neurons from the network. For example, \cite{alvarez2016learning} use a group sparsity regularization to reduce the width of the network. \cite{zoph2016neural} use reinforcement learning to search for the optimal architecture. Our proposed technique is related to these in the sense that we use an overcomplete network, and automatically adapt the effective depth of the network to learn an appropriate capacity network based on the data. Unlike other model selection techniques which work only in the batch learning setting, our method is designed for the online learning setting.

\section{Online Deep Learning}

\subsection{Problem Setting}

Without loss of generality, consider an online classification task. The goal of online deep learning is to learn a function $\F:\mathbb{R}^d \rightarrow \mathbb{R}^C$ based on a sequence of training examples $\D=\{(\x_1,y_1),\ldots,(\x_T,y_T)\}$, that arrive sequentially, where $\x_t\in \mathbb{R}^d$ is a $d$-dimensional instance representing the features and $y_t \in \{0,1\}^C$ is the class label assigned to $\x_t$ and $C$ is the number of classes. The prediction is denoted by $\hat{y_t}$, and the performance of the learnt functions are usually evaluated based on the cumulative prediction error: $ \epsilon_T = \frac{1}{T}{\sum_{t=1}^T \I_{(\hat{y_t} \neq y_t)}},$ where $\I$ is the indicator function resulting in $1$ if the condition is true, and $0$ otherwise. To minimize the classification error over the sequence of $T$ instances, a loss function (e.g., squared loss, cross-entropy, etc.) is often chosen for minimization. In every online iteration, when an instance $\x_t$ is observed, and the model makes a prediction, the environment then reveals the ground truth of the class label, and finally the learner makes an update to the model (e.g., using online gradient descent).

\subsection{Backpropagation: Preliminaries and  Limitations}
For typical online learning algorithms, the prediction function $\F$ is either a linear or kernel-based model. In the case of Deep Neural Networks (DNN), it is a set of stacked linear transformations, each followed by a nonlinear activation. Given an input $\x \in \mathbb{R}^d$, the prediction function of DNN with $L$ hidden layers ($\h^{(1)}, \dots,\h^{(L)}$) is recursively given by:

\bq
\nonumber &\F(\x)& = \text{softmax}(W^{(L+1)} \h^{(L)}) \quad \text{where} \\ 
\nonumber &\h^{(l)}& = \sigma(W^{(l)} \h^{(l-1)})\ \quad \forall l = 1,\dots, L; \quad\h^{(0)} = \x
\eq

where $\sigma$ is an activation function, e.g., sigmoid, tanh, ReLU, etc. This equation represents a feedforward step of a neural network. The hidden layers $\h^{(l)}$ are the feature representations learnt during the training procedure. To train a model with such a configuration, we use the cross-entropy loss function denoted by $\mathcal{L}(\F(\x), y)$. We aim to estimate the optimal model parameters $W_i$ for $i = 1,\dots (L+1)$ by applying Online Gradient Descent (OGD) on this loss function. Following the online learning setting, the update of the model in each iteration by OGD is given by:
\bq
\nonumber
&W^{(l)}_{t+1} \leftarrow W^{(l)}_{t} - \eta \nabla_{W^{(l)}_t} \mathcal{L}(\F(\x_t), y_t) \quad \nonumber \forall l = 1, \dots, L+1
\eq
where $\eta$ is the learning rate. Using backpropagation, the chain rule of differentiation is applied to compute the gradient of the loss with respect to $W^{(l)}$ for $l \le L$. 

Unfortunately, using such a model for an online learning (i.e. Online Backpropagation) task faces several issues with convergence. Most notably: (i) For such models, a fixed depth of the neural network has to be decided a priori, and this cannot be changed during the training process. This is problematic, as determining the depth is a difficult task. Moreover, in an online setting, different depths may be suitable for a different number of instances to be processed, e.g. because of convergence reasons, shallow networks maybe preferred for small number of instances, and deeper networks for large number of instances. Our aim is to exploit the fast convergence of shallow networks at the initial stages, and the power of deep representation at a later stage; (ii) vanishing gradient is well noted problem that slows down learning.  This is even more critical in an online setting, where the model needs to make predictions and learn simultaneously; (iii) diminishing feature reuse, according to which many useful features are lost in the feedforward stage of the prediction. This is very critical for online learning, where it is imperative to quickly find the important features, to avoid poor performance for the initial training instances.

To address these issues, we design a training scheme for \emph{Online Deep Learning} through a Hedging strategy: Hedge Backpropagation (HBP). Specifically, HBP uses an over-complete network, and automatically decides how and when to adapt the depth of the network in an online manner.

\begin{figure} [h]
	\centering
	\includegraphics[width=0.4\textwidth]{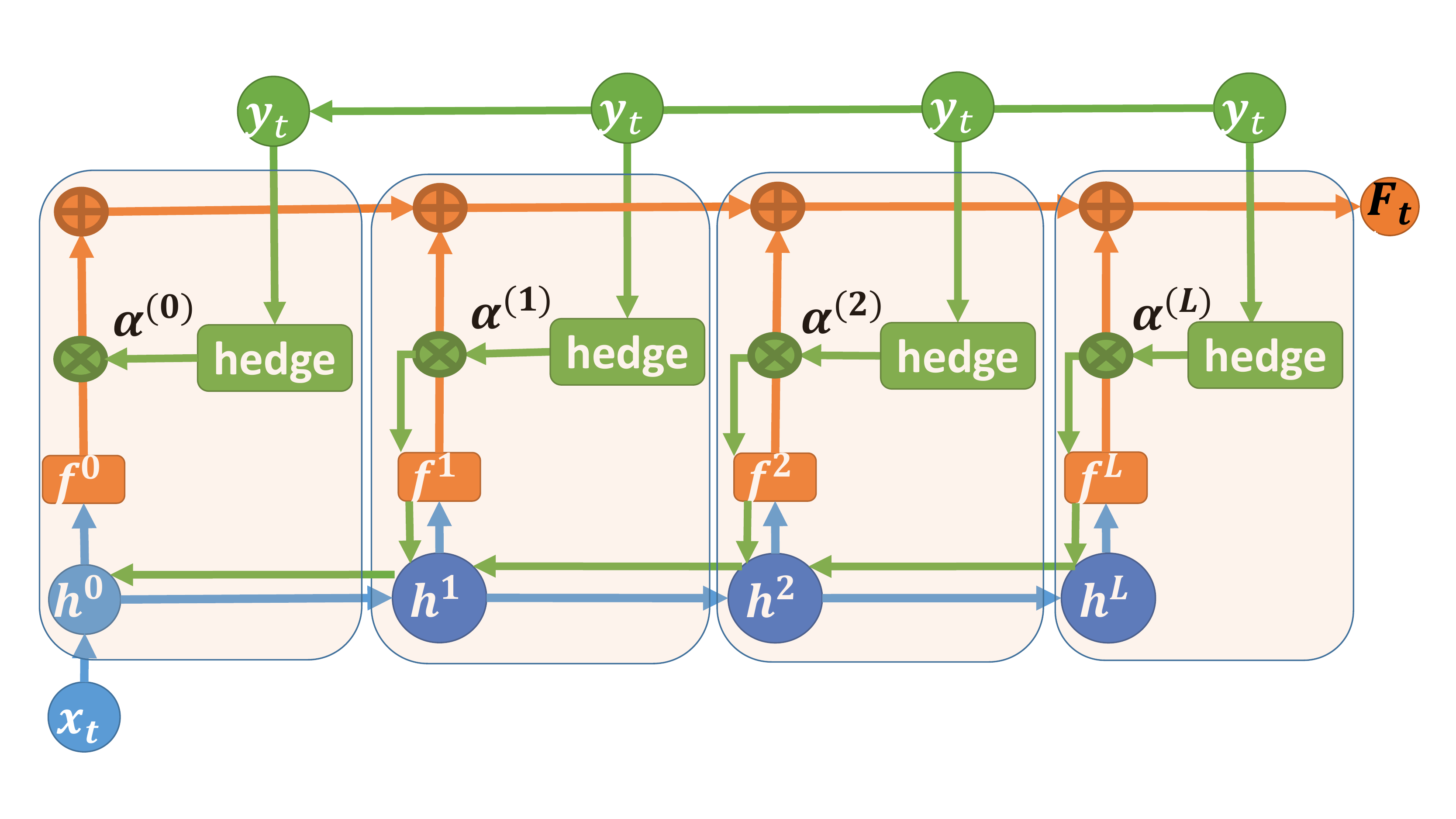}
	\caption{Online Deep Learning framework using Hedge Backpropagation (HBP). Blue lines represent feedforward flow for computing hidden layer features. Orange lines indicate softmax output followed by the hedging combination at prediction time. Green lines indicate the online updating flows with the hedge backpropagation approach.}
	\label{fig:hbp}
\end{figure}

\subsection{Hedge Backpropagation (HBP)}

Figure \ref{fig:hbp} illustrates the online deep learning framework, for training DNNs using  Hedge Backpropagation.

Consider a deep neural network with $L$ hidden layers (i.e. the maximum capacity of the network that can be learnt is one with $L$ hidden layers), the prediction function for the proposed Hedged Deep Neural Network is given by:
\bq
\label{eq:hdnn}
\F(\x) & = & \sum_{l=0}^L \alpha^{(l)} \f^{(l)}\quad \text{where}\\
\nonumber  \f^{(l)}   &=& \mathrm{softmax}(\h^{(l)} \Theta^{(l)}),\ \forall l = 0,\dots, L\\
\nonumber   \h^{(l)}  & = & \sigma(W^{(l)} \h^{(l-1)}),\ \forall l = 1,\dots, L\\
\nonumber   \h^{(0)}  & = & \x
\eq
Here we have designed a new architecture, and introduced two sets of new parameters $\Theta^{(l)}$ (parameters for $\f^{(l)}$) and $\alpha$, that have to be learnt. Unlike the original network, in which the final prediction is given by a classifier using the feature representation $\h^{(L)}$, here the prediction is weighted combination of classifiers learnt using the feature representations from $\h^{(0)}, \dots, \h^{(L)}$. Each classifier $\f^{(l)}$ is parameterized by $\Theta^{(l)}$. Note that there are a total of $L+1$ classifiers. The final prediction of this model is a weighted combination of the predictions of all  classifiers, where the weight of each classifier is denoted by $\alpha^{(l)}$, and the loss suffered by the model is $\mathcal{L}(\F(\x), y) = \sum_{l=0}^L \alpha^{(l)}\mathcal{L}(\f^{(l)}(\x), y)$.
During the online learning procedure, we need to learn $\alpha^{(l)}$, $\Theta^{(l)}$ and $W^{(l)}$.

We propose to learn $\alpha^{(l)}$ using the Hedge Algorithm \cite{Freund1997}. At the first iteration, all  weights $\alpha$ are uniformly distributed, i.e., $\alpha^{(l)} = \frac{1}{L+1}, l = 0, \dots, L$. At every iteration, the classifier $\f^{(l)}$ makes a prediction $\hat{y_t}^{(l)}$. When the ground truth is revealed, the classifier's weight is updated based on the loss suffered by the classifier as: 
\bq 
\nonumber \alpha^{(l)}_{t+1} \leftarrow \alpha^{(l)}_t \beta ^{\mathcal{L}(\f^{(l)}(\x), y)}
\eq

where $\beta \in (0,1)$ is the discount rate parameter, and $\mathcal{L}(\f^{(l)}(\x), y) \in (0,1)$ \cite{Freund1997}. Thus, a classifier's weight is discounted by a factor of $\beta^{\mathcal{L}(\f^{(l)}(\x), y)}$ in every iteration. At the end of every round, the weights $\alpha$ are normalized such that $\sum_l \alpha^{(l)}_t = 1$.

Learning the parameters $\Theta^{(l)}$ for all the classifiers can be done via online gradient descent \cite{Zinkevich2003}, where the input to the $l^{th}$ classifier is $\h^{(l)}$. This is similar to the update of the weights of the output layer in the original feedforward networks. This update is given by:
\bq
\label{eq:classifierUpdate}
\Theta^{(l)}_{t+1}  &\leftarrow&  \Theta^{(l)}_t - \eta \nabla_{\Theta^{(l)}_t} \mathcal{L}(\F(\x_t, y_t)) \\
\nonumber
&=&  \Theta^{(l)}_t - \eta \alpha^{(l)} \nabla_{\Theta^{(l)}_t} \mathcal{L}(\f^{(l)}, y_t)
\eq

Updating the feature representation parameters $W^{(l)}$ is more tricky. Unlike the original backpropagation scheme, where the error derivatives are backpropagated from the output layer, here, the error derivatives are backpropagated from \emph{every} classifier $\f^{(l)}$. Thus, using the adaptive loss function $\mathcal{L}(\F(\x), y) = \sum_{l=0}^L \alpha^{(l)}\mathcal{L}(\f^{(l)}(\x), y)$ and applying OGD rule, the update rule for $W^{(l)}$ is given by:
\bq
\label{eq:hdnnBackprop}
W^{(l)}_{t+1}  \leftarrow W^{(l)}_{t} - \eta \sum_{j=l}^L \alpha^{(j)} \nabla _{W^{(l)}} \mathcal{L}(\f^{(j)}, y_t)
\eq
where $\nabla _{W^{(l)}} \mathcal{L}(\f^{(j)}, y_t)$ is computed via backpropagation from error derivatives of $\f^{(j)}$. Note that the summation (in the gradient term) starts at $j=l$ because the shallower classifiers do not depend on $W^{(l)}$ for making predictions. In effect, we are computing the gradient of the final prediction with respect to the backpropagated derivatives of a predictor at every depth weighted by $\alpha^{(l)}$ (which is an indicator of the performance of the classifier). Hedge enjoys a regret of $R_T \le \sqrt{T \ln N}$, where N is the number of experts \cite{freund1999adaptive}, which in our case is the network depth. This gives an effective model selection approach to adapt to the optimal network depth automatically online.

Based on the intuition that shallower models tend to converge faster than deeper models \cite{chen2015net2net,larsson2016fractalnet,gulcehre2016mollifying}, using a hedging strategy would lower $\alpha$ weights of deeper classifiers to a very small value (due to poor initial performance as compared to shallower classifiers), which would  affect the update in Eq. \eqref{eq:hdnnBackprop}, and result in deeper classifiers having slow learning. To alleviate this, we introduce a \emph{smoothing} parameter $s \in (0,1)$ which is used to set a minimum weight for each classifier. After the weight update of the classifiers in each iteration, the weights are set as:
$ \alpha^{(l)} \leftarrow \max\Big(\alpha^{(l)}, \frac{s}{L}\Big)$
This maintains a minimum weight for a classifier of every depth and helps us achieve a tradeoff between exploration and exploitation. $s$ encourages all classifiers at every depth to affect the backprop update (exploring high capacity deep classifiers, and enabling deep classifiers to perform as good as shallow ones),  while hedging the model exploits the best performing classifier. Similar strategies have been used in Multi-arm bandit setting, and online learning with expert advice to trade off exploration and exploitation \cite{auer2002nonstochastic,Hoi2013}. Algorithm \ref{alg1} outlines ODL using HBP.

\begin{algorithm}[htp]
	\caption{Online Deep Learning (ODL) using HBP}\label{alg1}
	INPUTS: Discounting Parameter: $\beta \in (0,1)$;\\
	Learning rate Parameter: $\eta$;  Smoothing Parameter: $s$
	
	\textbf{Initialize}: {$\F(\x) = $ DNN with $L$ hidden layers and $L+1$ classifiers $\f^{(l)}, \forall l = 0, \dots, L$; $\alpha^{(l)} = \frac{1}{L+1}, \forall l = 0, \dots, L $}
	
	\begin{algorithmic}
		\FOR{t = 1,\dots,T}
		\STATE Receive instance: $\x_t$
		\STATE Predict $\hat{y_t} = \F_t(\x_t) = \sum_{l=0}^L \alpha_t^{(l)} \f_t^{(l)} $ as per Eq. \eqref{eq:hdnn}
		\STATE Reveal true value $y_t$
		
		\STATE Set $\mathcal{L}^{(l)}_t = \mathcal{L}(\f^{(l)}_t(\x_t),y_t), \forall l, \dots, L$;
		\STATE Update $\Theta^{(l)}_{t+1}, \forall l = 0, \dots, L$  as per Eq. \eqref{eq:classifierUpdate};
		\STATE Update $W^{(l)}_{t+1}, \forall l = 1, \dots, L$ as per Eq. \eqref{eq:hdnnBackprop};
		\STATE Update $\alpha^{(l)}_{t+1} = \alpha^{(l)}_{t} \beta^{\mathcal{L}^{(l)}_t}, \forall l = 0, \dots, L$;
		\STATE Smoothing $\alpha^{(l)}_{t+1} = \max(\alpha^{(l)}_{t+1}, \frac{s}{L}), \forall l = 0, \dots, L$ ;
		\STATE Normalize $\alpha^{(l)}_{t+1} = \frac{\alpha^{(l)}_{t+1}}{Z_t}$ where $Z_t = \sum\limits_{l=0}^L\alpha^{(l)}_{t+1}$
		\ENDFOR
	\end{algorithmic}
\end{algorithm}


\subsection{Discussion}

HBP has the following properties:
(i) it identifies a neural network of an appropriate depth based on the performance of the classifier at that depth. This is a form of Online Learning with Expert Advice,\cite{Cesa-Bianchi2006}, where the experts are DNNs of varying depth, making the DNN robust to depth.
(ii) it offers a good initialization for deeper networks, which are encouraged to match the performance of shallower networks (if unable to beat them). This facilitates knowledge transfer from shallow to deeper networks (\cite{chen2015net2net,wei2016network}), thus simulating student-teacher learning;
(iii) it makes the learning robust to vanishing gradient and diminishing feature reuse by using a multi-depth architecture where gradients are backpropagated from shallower classifiers, \emph{and} the low level features are used for the final prediction (by hedge weighted prediction);
(iv) it can be viewed as an ensemble of multi-depth networks which are competing and collaborating to improve the final prediction performance. The competition is induced by Hedging , whereas the collaboration is induced by sharing feature representations;
(v) This allows our algorithms to continuously learn and adapt as and when it sees more data, enabling a form of life-long learning \cite{lee2016dual}; 
(vi) In concept drifting scenarios \cite{gamajo2014survey}, traditional online backpropagation would suffer from slow convergence for deep networks (when the concepts would change), whereas, HBP is able to adapt quickly due to hedging; and
(vii) While HBP could be trivially adapted to Convolutional Networks for computer vision tasks, these tasks typically have many classes with few instances per class, which makes it hard to obtain good results in just one-pass through the data (online setting). Thus our focus is on pure online settings  where a large number of instances arrive in a stream and exhibit complex nonlinear patterns.

\section{Experiments}

\subsection{Datasets}

We consider several large scale datasets. Higgs and Susy are Physics datasets from UCI repository. For Higgs, we sampled 5 million instances. We generated 5 million instances from Infinite MNIST generator \cite{loosli2007training}. We also evaluated on 3 synthetic datasets. The first (Syn8) is generated from a randomly initialized DNN comprising 8-hidden layers (of width 100 each). The other two are concept drift datasets CD1 and CD2. In CD1, 2 concepts (C1 and C2), appear in the form C1-C2-C1, with each segment comprising a third of the data stream. Both C1 and C2 were generated from a 8-hidden layer network. CD2 has 3 concepts with C1-C2-C3, where C1 and C3 are generated from a 8-hidden layer network, and C2 from a shallower 6-hidden layer network.
Other details are in Table \ref{tab:data}.

\begin{table}[h!]
	\small
	\centering
	\caption{Datasets}
	\begin{tabular}{|c|c|c|c|}
		\hline
		Data  & \#Features & \#Instances & Type \\
		\hline
		Higgs & 28    & 5m    & Stationary \\
		Susy  & 18    & 5m    & Stationary \\
		i-mnist & 784 & 5m    & Stationary \\
		Syn8  & 50    & 5m    & Stationary \\
		CD1   & 50    & 4.5m  & Concept Drift \\
		CD2   & 50    & 4.5m  & Concept Drift \\
		\hline
	\end{tabular}%
	\label{tab:data}%
\end{table}

\subsection{Limitations of traditional Online BP: Difficulty in Depth Selection prior to Training}

First we demonstrate the difficulty of DNN model selection in the online setting. We compare the error rate of DNNs of varying depth, in different segments of the data. All models were trained online, and we evaluate their performance in different windows (or stages) of the learning process. See Table \ref{tab:dnnDepth}. In the first 0.5\% of the data, the shallowest network obtains the best performance indicating faster convergence - which would indicate that we should use the shallow network for the task. In the segment from [10-15]\%, a 4-layer DNN seems to have the best performance in most cases. And in the segment from [60-80]\% of the data, an 8-layer network gives a better performance. This indicates that deeper networks took a longer time to converge, but at a later stage gave a better performance. Looking at the final error, it does not give us conclusive evidence of what depth of network would be the most suitable. Furthermore, if the datastream had more instances, a deeper network may have given an overall better performance. This demonstrates the difficulty in model selection for learning DNNs online, where typical validation techniques are ineffective. Ideally we want to exploit the convergence of the shallow DNNs in the beginning and the power of deeper representations later.
\begin{table*}
	\small
	\centering
	\caption{Online error rate of DNNs of varying Depth. All models were trained online $t=1,\dots,T$. We evaluate the performance in different segments of the data. L is the number of layers in the DNN.}
	
	\begin{tabular}{|c|ccc|ccc|ccc|ccc|}
		\hline
		& \multicolumn{3}{c|}{\textbf{Final Cumulative Error }} & \multicolumn{3}{c|}{\textbf{Segment [0-0.5]\% Error}} & \multicolumn{3}{c|}{\textbf{Segment [10-15]\% Error}} & \multicolumn{3}{c|}{\textbf{Segment [60-80]\% Error}} \\
		\hline
		\textbf{L}& \textbf{Higgs} & \textbf{Susy} & \textbf{Syn8} & \textbf{Higgs} & \textbf{Susy} & \textbf{Syn8} & \textbf{Higgs} & \textbf{Susy} & \textbf{Syn8} & \textbf{Higgs} & \textbf{Susy} & \textbf{Syn8} \\
		\hline
		\textbf{3} & 0.2724 & 0.2016 & 0.3936 & \textbf{0.3584} & \textbf{0.2152} & \textbf{0.4269} & 0.2797 & \textbf{0.2029} & 0.4002 & 0.2668 & 0.2004 & 0.3901 \\
		\textbf{4} & 0.2688 & \textbf{0.2014} & \textbf{0.3920} & 0.3721 & 0.2197 & 0.4339 & \textbf{0.2775} & \textbf{0.2030} & \textbf{0.3989} & 0.2617 & 0.2004 & \textbf{0.3876} \\
		\textbf{8} & \textbf{0.2682} & 0.2016 & 0.3936 & 0.3808 & 0.2218 & 0.4522 & 0.2794 & 0.2036 & 0.4018 & \textbf{0.2613} & \textbf{0.1997} & 0.3888 \\
		\textbf{16} & 0.2731 & 0.2037 & 0.4025 & 0.4550 & 0.2312 & 0.4721 & 0.2831 & 0.2050 & 0.4121 & 0.2642 & 0.2027 & 0.3923 \\
		\hline
	\end{tabular}%
	
	\label{tab:dnnDepth}%
\end{table*}%

\subsection{Baselines}
We aim to learn a 20 layer DNN in the online setting, with 100 units in each hidden layer. As baselines, we learn the 20 layer network online using OGD (Online Backpropagation), OGD Momentum, OGD Nesterov, and Highway Networks. Since a 20 layer network would be very difficult to learn in the online setting, we also compare the performance of shallower networks --- DNNs with fewer layers (2,3,4,8,16), trained using Online BP. We used ReLU Activation, and a fixed learning rate of 0.01 (chosen, as it gave the best final error rate for all DNN-based baselines). For momentum techniques, a fixed learning rate of 0.001 was used, and we finetuned the momentum parameter to obtain the best performance for the baselines. For HBP, we attached a classifier to each of the 19 hidden layers (and not directly to the input). This gave 19 classifiers each with depth from $2,\dots,20$. We set $\beta = 0.99$ and the smoothing parameter $s=0.2$. Implementation was in Keras \cite{chollet2015keras}. We also compared with representative state of the art linear online learning algorithms (OGD, Adaptive Regularization of Weights (AROW), Soft-Confidence Weighted Learning (SCW) \cite{hoi2014libol}) and kernel online learning algorithms (Fourier Online Gradient Descent (FOGD) and Nystrom Online Gradient Descent (NOGD)\cite{lu2015large}) .

\subsection{Evaluation of Online Deep Learning Algorithms}

The final cumulative error obtained by all the baselines and the proposed HBP technique can be seen in Table \ref{tab:finalError}. First, traditional online learning algorithms (linear and kernel) have relatively poor performance on complex datasets.
Next, in learning with a 20-layer network, the convergence is slow, resulting in poor overall performance. While second order methods utilizing momentum and highway networks are able to offer some advantage over simple Online Gradient Descent, they can be easily beaten by a relatively shallower networks in the online setting.
We observed before that relatively shallower networks could give a competitive performance in the online setting, but lacked the ability to exploit the power of depth at a later stage. In contrast, HBP enjoyed the best of both worlds, by allowing for faster convergence initially, and making use of the power of depth at a later stage. This way HBP was able to do automatic model selection online, enjoying merits of both shallow and deep networks, and this resulted in HBP outperforming all the DNNs of different depths, in terms of online performance. It should be noted that the optimal depth for DNN is not known before the learning process, and even then HBP outperforms all DNNs at any depth.

\begin{figure*}
	\centering
	\subfigure[HIGGS]{
		\includegraphics[width=.32\textwidth,height=4cm]{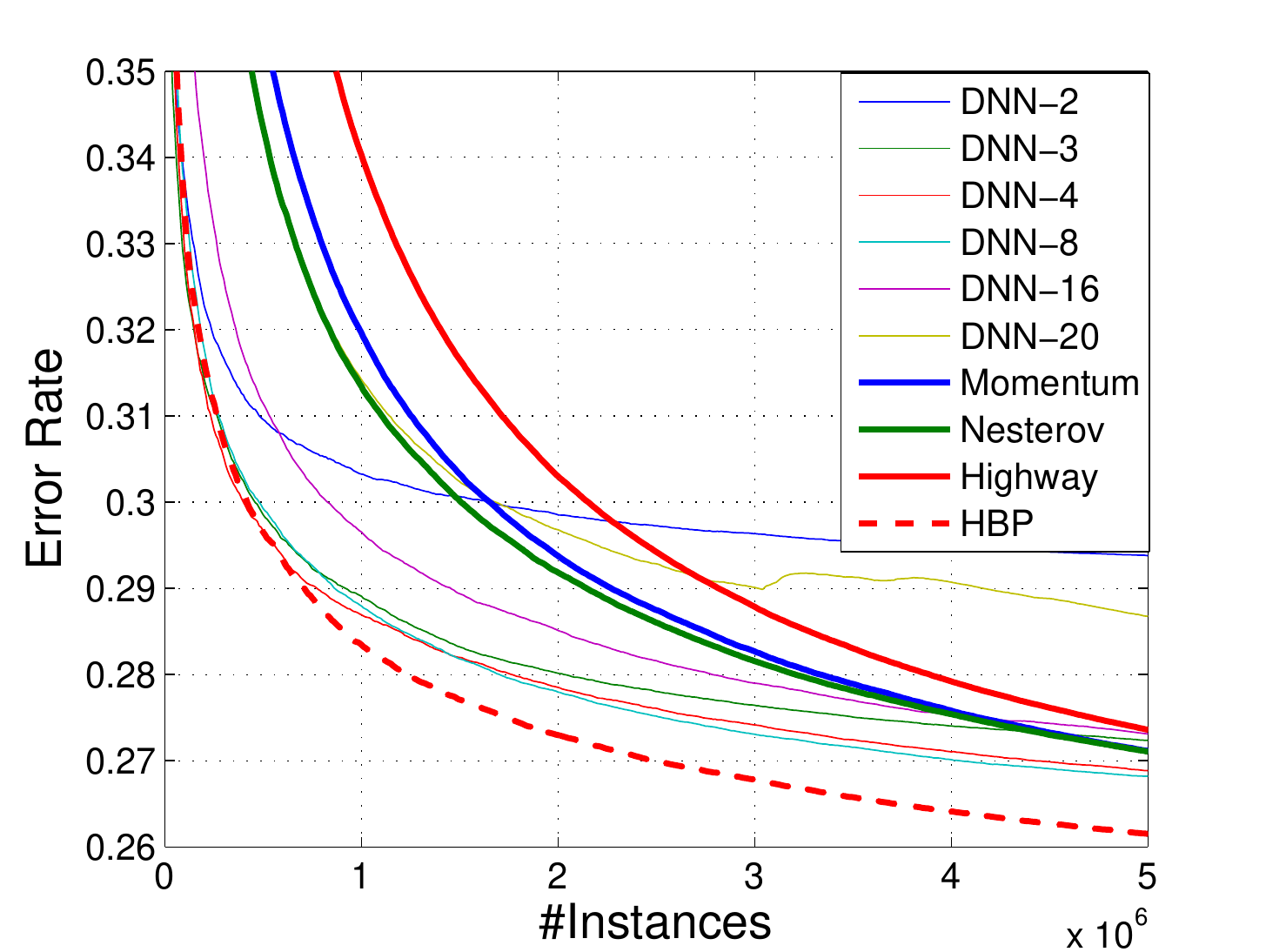}}
	\subfigure[SUSY]{
		\includegraphics[width=.32\textwidth,height=4cm]{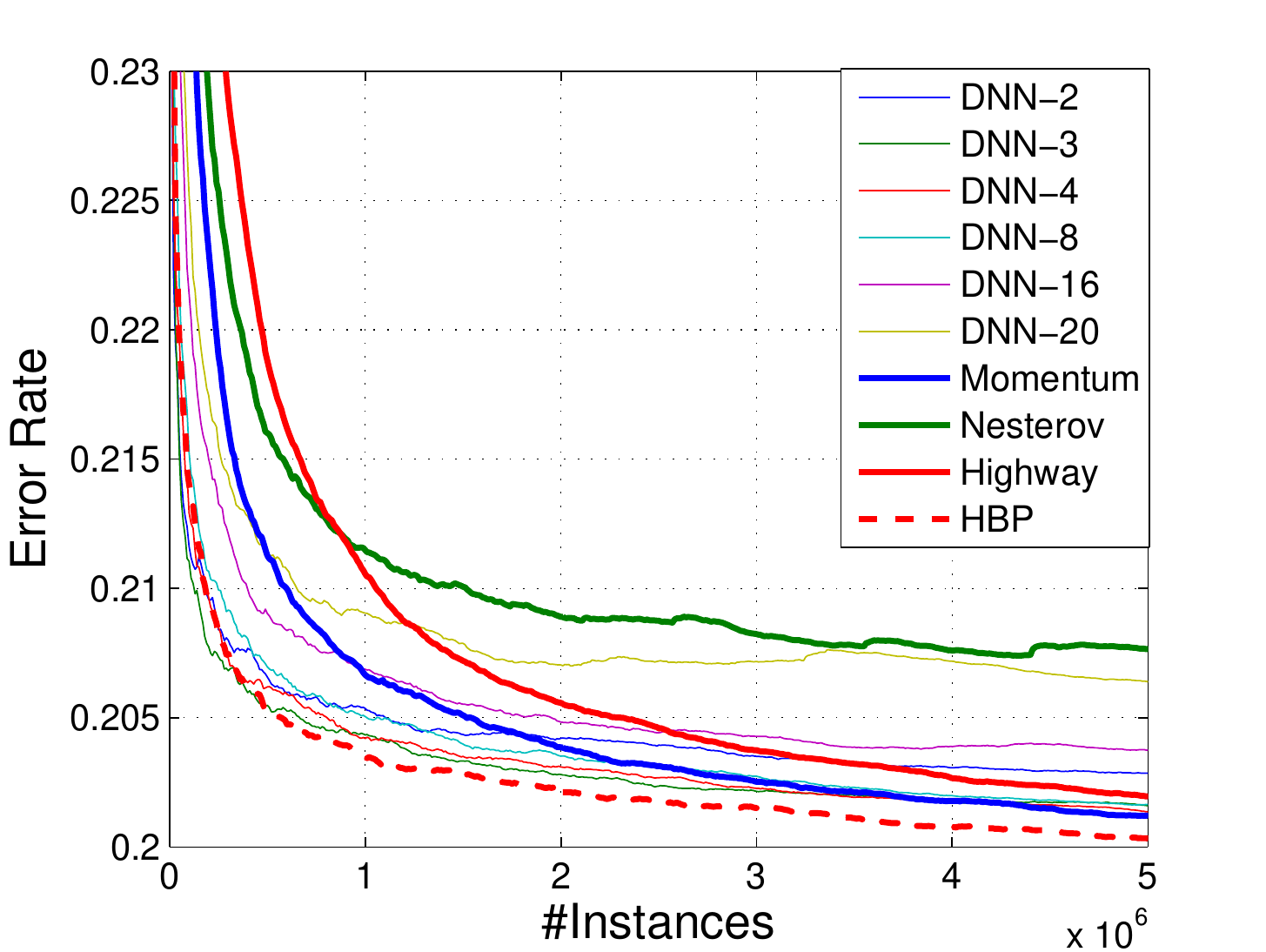}}
	\subfigure[Inf-MNIST (i-mnist)]{
		\includegraphics[width=.32\textwidth,height=4cm]{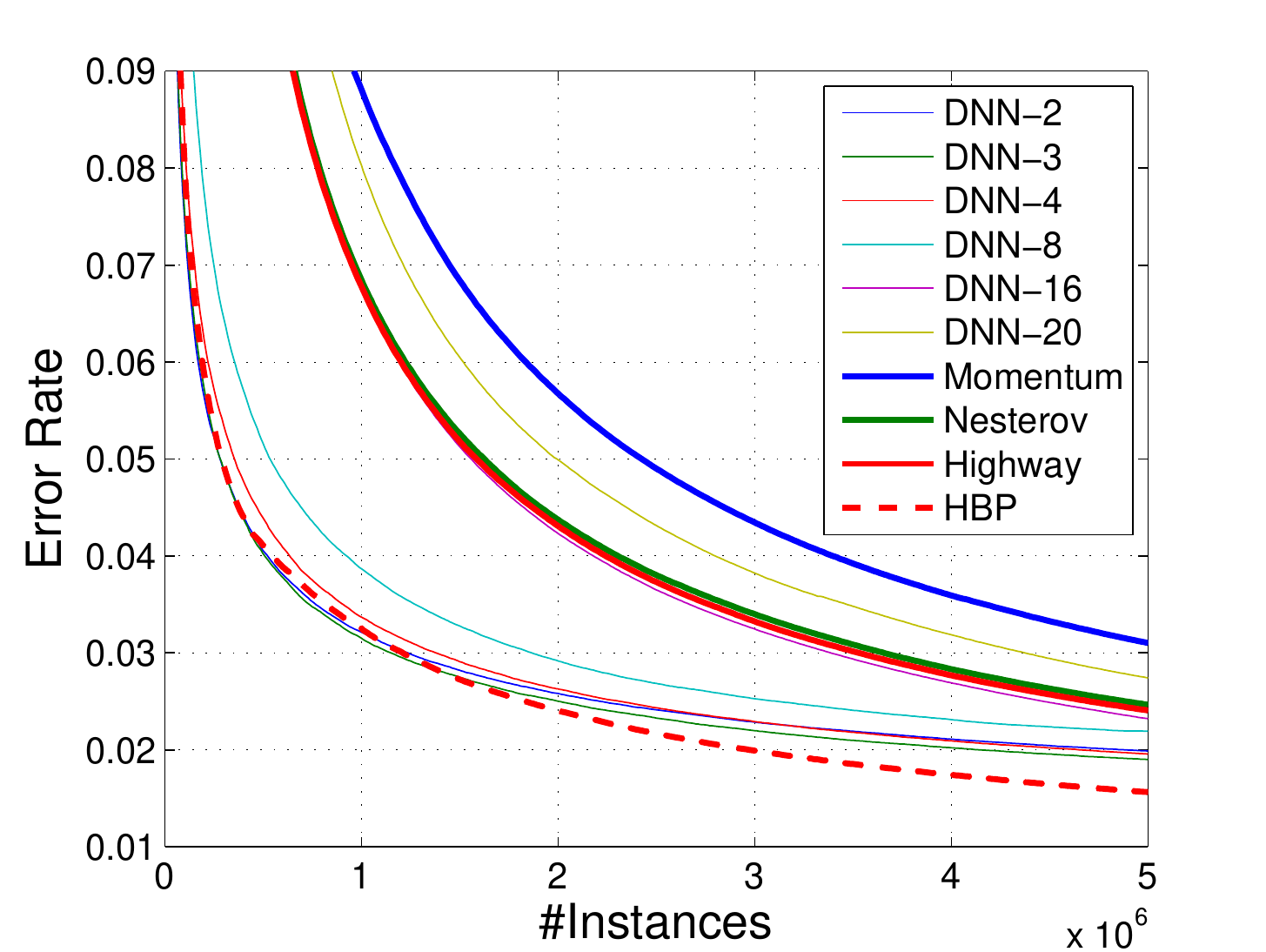}}
	\subfigure[SYN8]{
		\includegraphics[width=.32\textwidth,height=4cm]{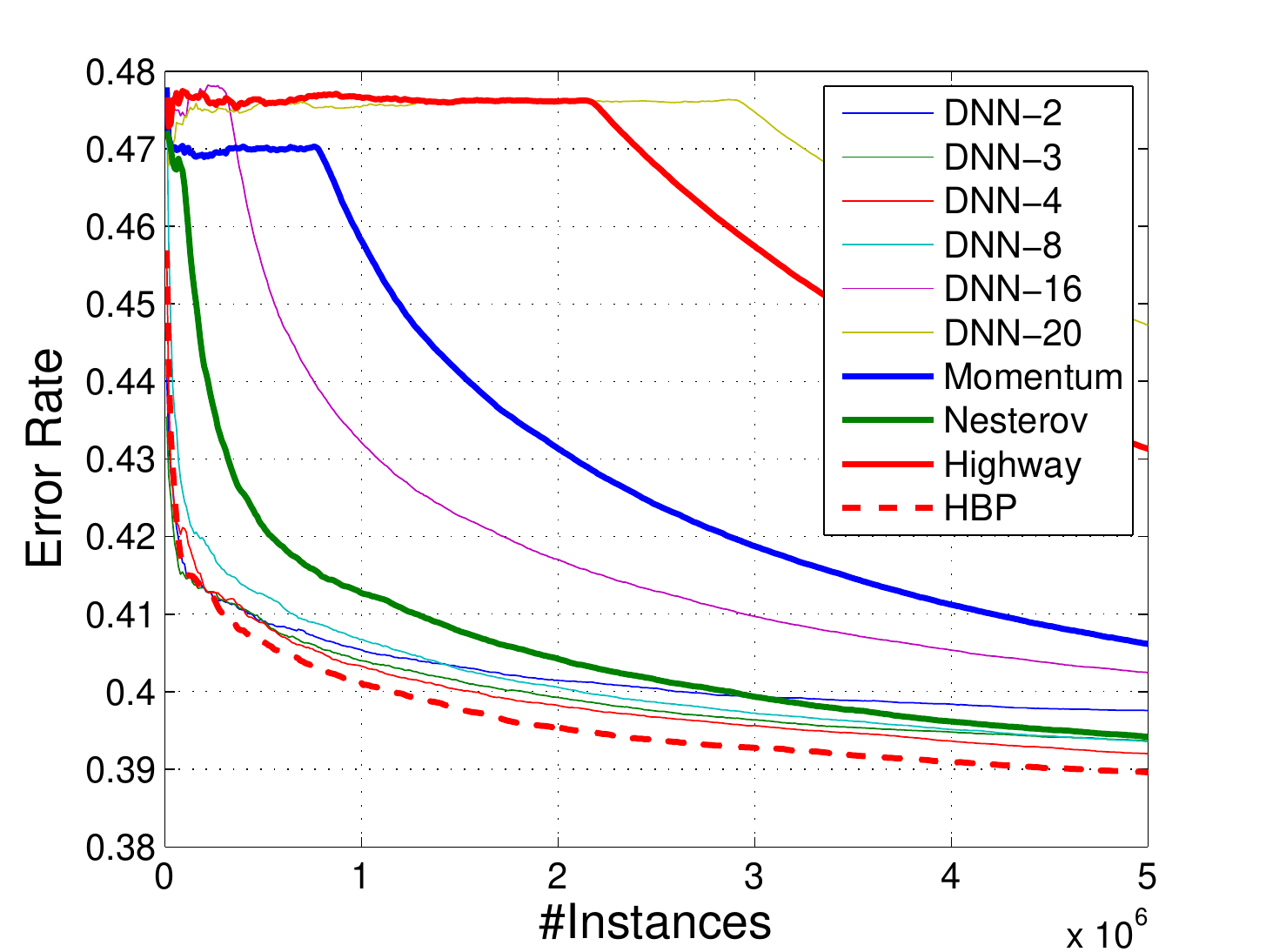}}
	\subfigure[Concept Drift 1 (CD1)]{
		\includegraphics[width=.32\textwidth,height=4cm]{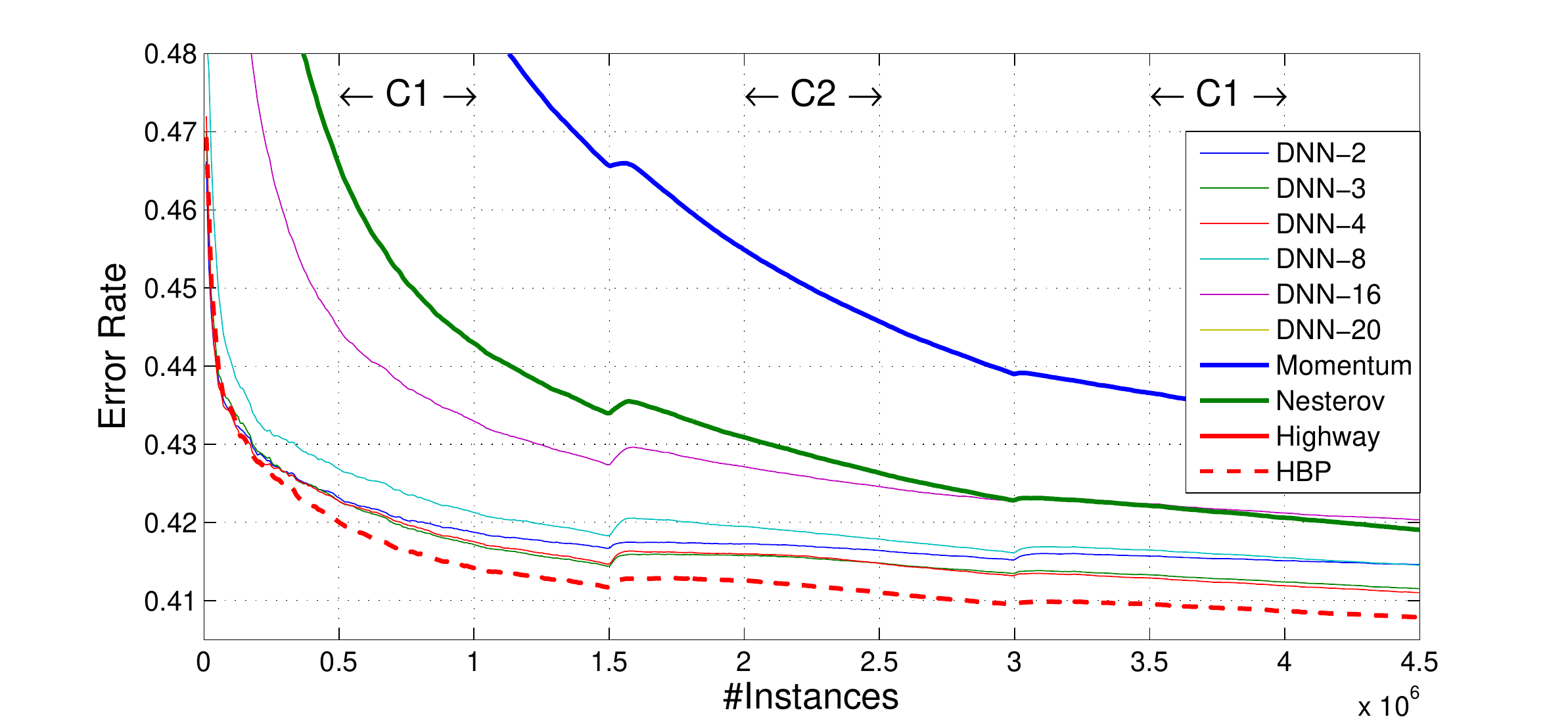}}
	\subfigure[Concept Drift 2 (CD2)]{
		\includegraphics[width=.32\textwidth,height=4cm]{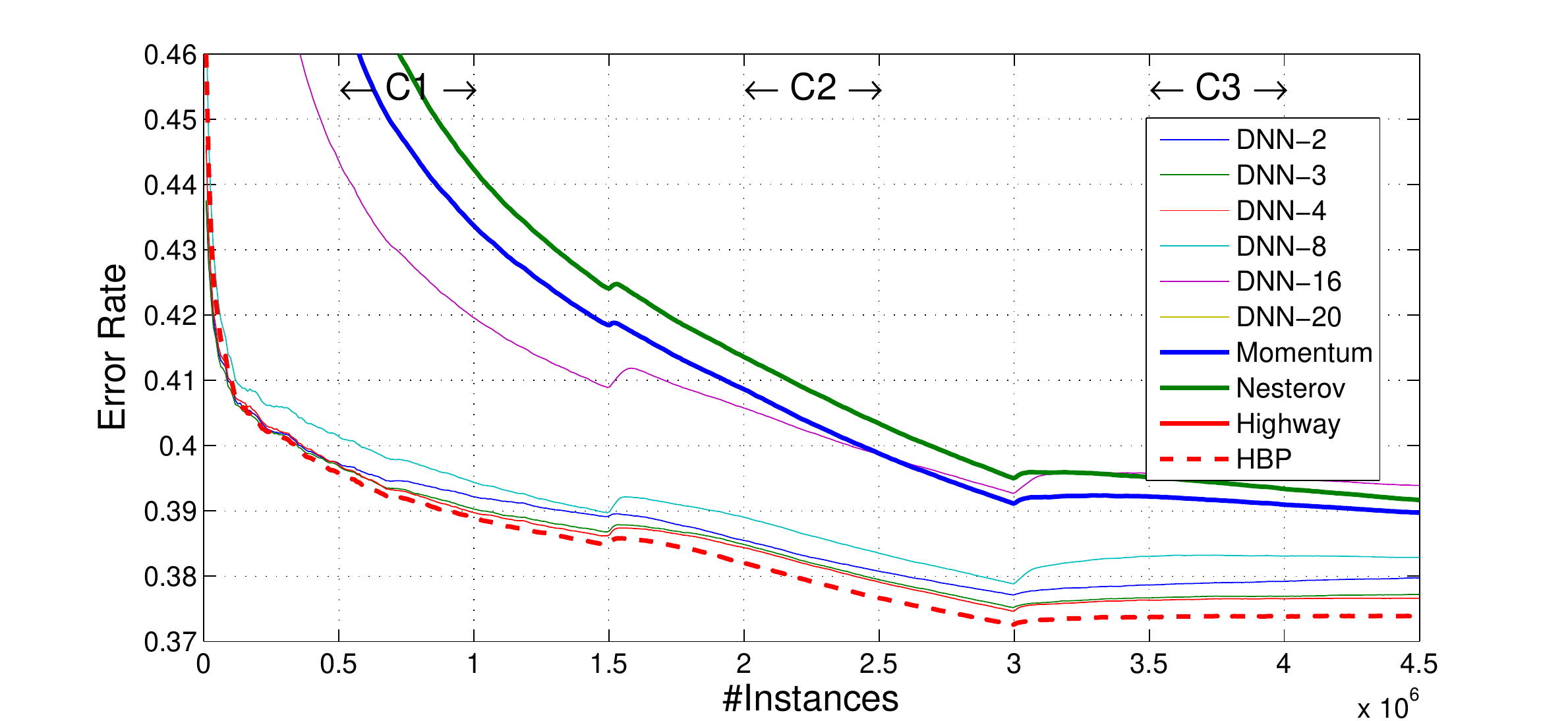}}
	\caption{Convergence behavior of DNNs in Online Setting on stationary and concept drifting data.}
	\label{fig:convergence}
\end{figure*}

In Figure \ref{fig:convergence}, we can see the convergence behavior of all the algorithms on the stationary as well as concept drift datasets. In the stationary datasets, HBP shows consistent outperformance over all the baselines. The only exception is in the very initial stages of the online learning phase, where shallower baselines are able to outperform HBP. This is not surprising, as HBP has many more parameters to learn. However, HBP is able to quickly outperform the shallow networks. The performance of HBP in concept drifting scenarios demonstrates its ability to adapt to the change fairly quickly, enabling usage of DNNs in the concept drifting scenarios. Looking at the performance of simple 20-layer (and 16-layer) networks on concept drifting data, we can see difficulty in utilizing deep representation for such scenarios.

\begin{table*}
	\centering
	\caption{Final Online Cumulative Error Rate obtained by the algorithms. Best performance is in bold.}
	\begin{tabular}{|c|c|c|cccccc|}
		\hline
		\textbf{Model} & \textbf{Method} & \textbf{Layers} & \textbf{Higgs} & \textbf{Susy} & \textbf{i-mnist} & \textbf{Syn8} & \textbf{CD1} & \textbf{CD2} \\
		\hline
		\hline
		\multirow{3}[0]{*}{\textbf{Linear OL}} & \textbf{OGD} & 1     & 0.3620 & 0.2160 &  0.1230     & 0.4070 & 0.4360 & 0.4270 \\
		& \textbf{AROW} & 1     & 0.3630 & 0.2160 &  0.1250     & 0.4050 & 0.4340 & 0.4250 \\
		& \textbf{SCW} & 1     & 0.3530 & 0.2140 &   0.1230    & 0.4050 & 0.4340 & 0.4250 \\
		\hline
		\hline
		\multirow{2}[0]{*}{\textbf{Kernel OL}} & \textbf{FOGD} & 2     & 0.2973 & 0.2021 &  0.0495     & 0.3962 & 0.4329 & 0.4193 \\
		& \textbf{NOGD} & 2     & 0.3488 & 0.2045 &  0.1045     & 0.4146 & 0.4455 & 0.4356 \\
		\hline
		\hline
		\multirow{10}[0]{*}{\textbf{DNNs}} & \textbf{OGD (Online BP)} & 2     & 0.2938 & 0.2028 & 0.0199 & 0.3976 & 0.4146 & 0.3797 \\
		& \textbf{OGD (Online BP)} & 3     & 0.2724 & 0.2016 & 0.0190 & 0.3936 & 0.4115 & 0.3772 \\
		& \textbf{OGD (Online BP)} & 4     & 0.2688 & 0.2014 & 0.0196 & 0.3920 & 0.4110 & 0.3766 \\
		& \textbf{OGD (Online BP)} & 8     & 0.2682 & 0.2016 & 0.0219 & 0.3936 & 0.4145 & 0.3829 \\
		& \textbf{OGD (Online BP)} & 16    & 0.2731 & 0.2037 & 0.0232 & 0.4025 & 0.4204 & 0.3939 \\
		& \textbf{OGD (Online BP)} & 20    & 0.2868 & 0.2064 & 0.0274 & 0.4472 & 0.4928 & 0.4925 \\
		& \textbf{OGD+Momentum} & 20    & 0.2711 & 0.2012 & 0.0310 & 0.4062 & 0.4312 & 0.3897 \\
		& \textbf{OGD+Nesterov} & 20    & 0.2711 & 0.2076 & 0.0247 & 0.3942 & 0.4191 & 0.3917 \\
		& \textbf{Highway} & 20    & 0.2736 & 0.2019 & 0.0241 & 0.4313 & 0.4928 & 0.4925 \\
		\cline{2-9}
		& \textbf{Hedge BP (proposed)} & 20    & \textbf{0.2615} & \textbf{0.2003} & \textbf{0.0156} & \textbf{0.3896} & \textbf{0.4079} & \textbf{0.3739} \\
		\hline
	\end{tabular}%
	\label{tab:finalError}%
\end{table*}%

\begin{figure*}
	\centering
	\subfigure[First 0.5\% of Data]{
		\includegraphics[width=.32\textwidth,height=3cm]{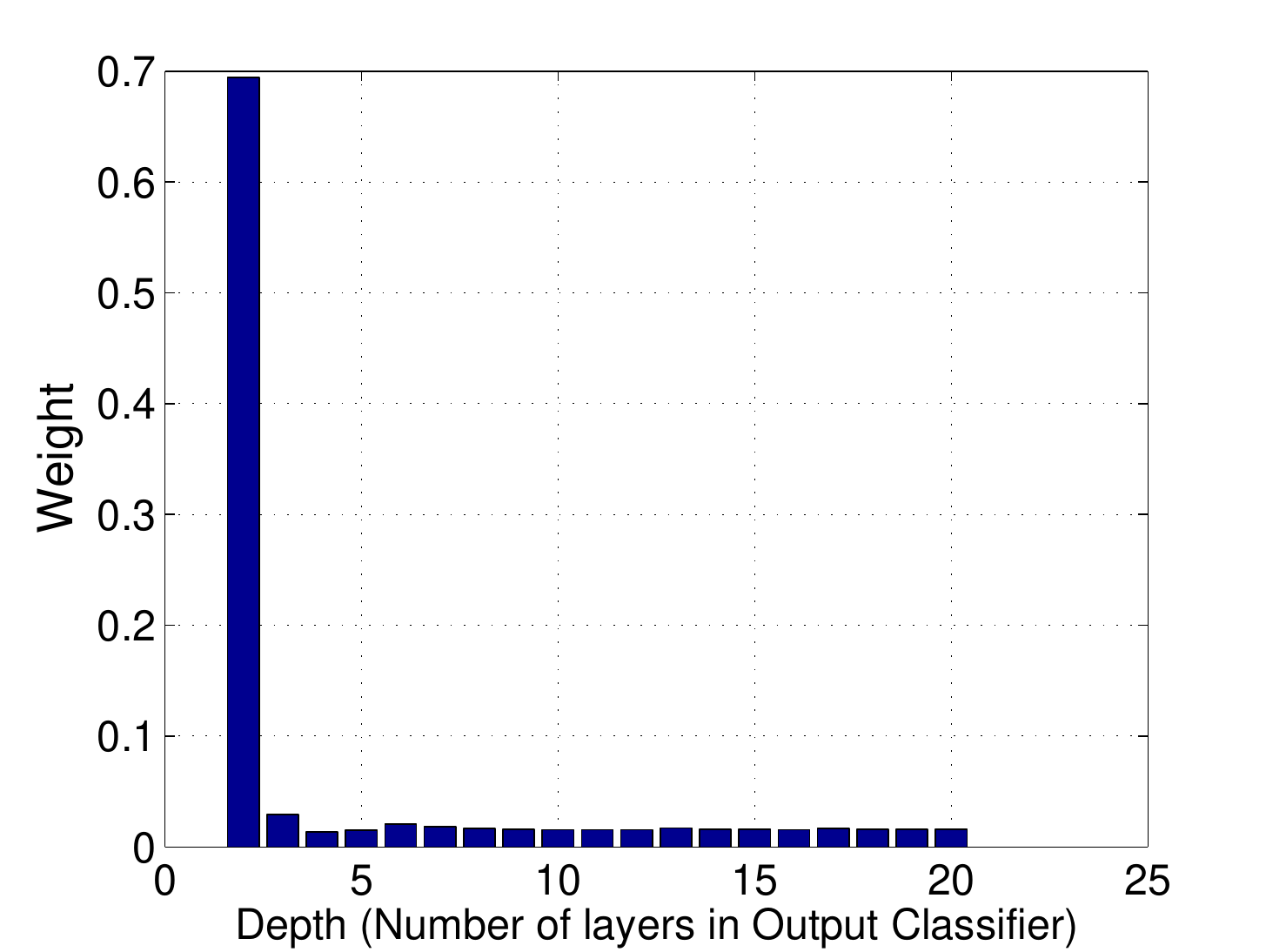}}
	\subfigure[10-15\% of Data]{
		\includegraphics[width=.32\textwidth,height=3cm]{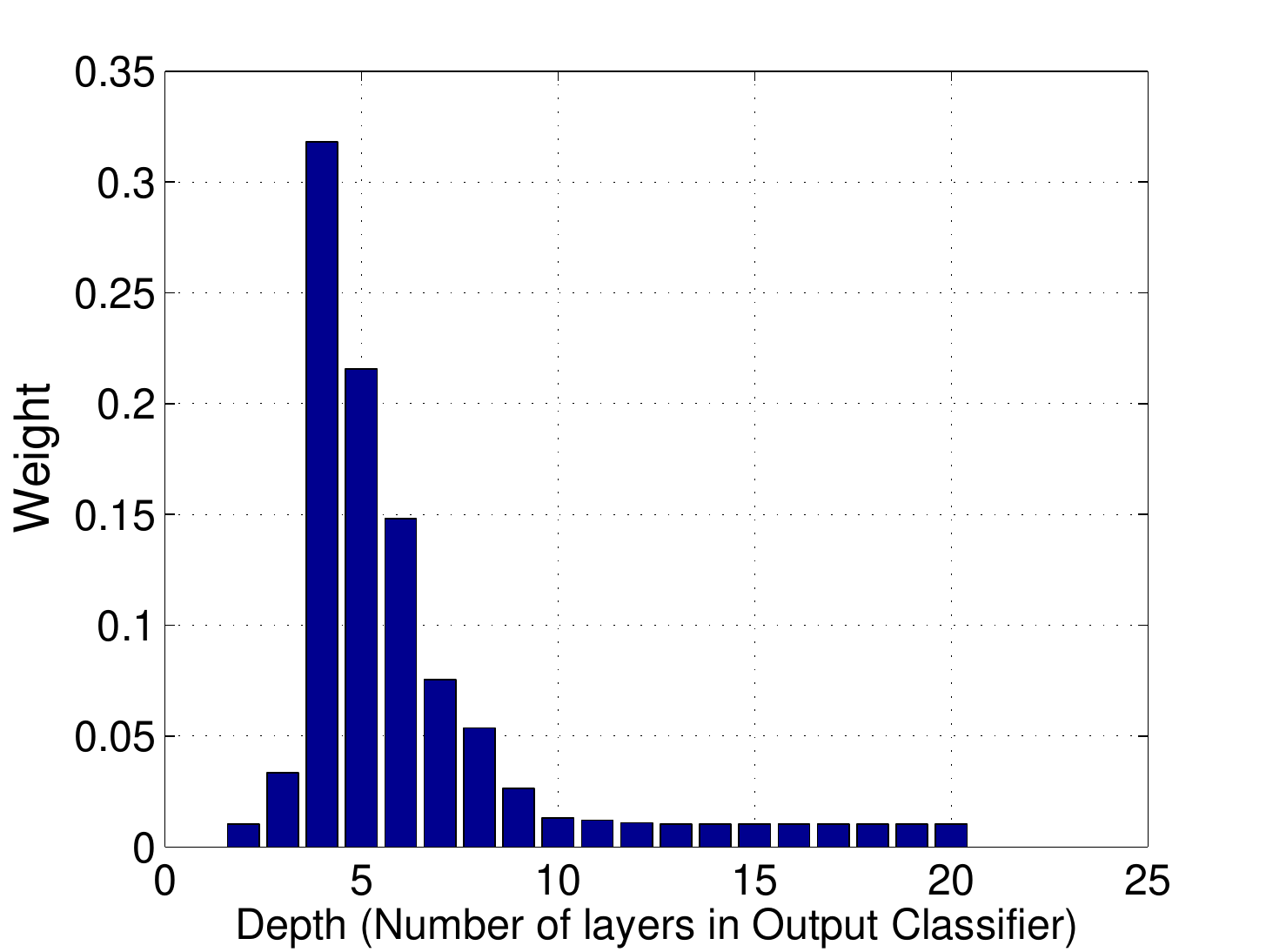}}
	\subfigure[60-80\% of Data]{
		\includegraphics[width=.32\textwidth,height=3cm]{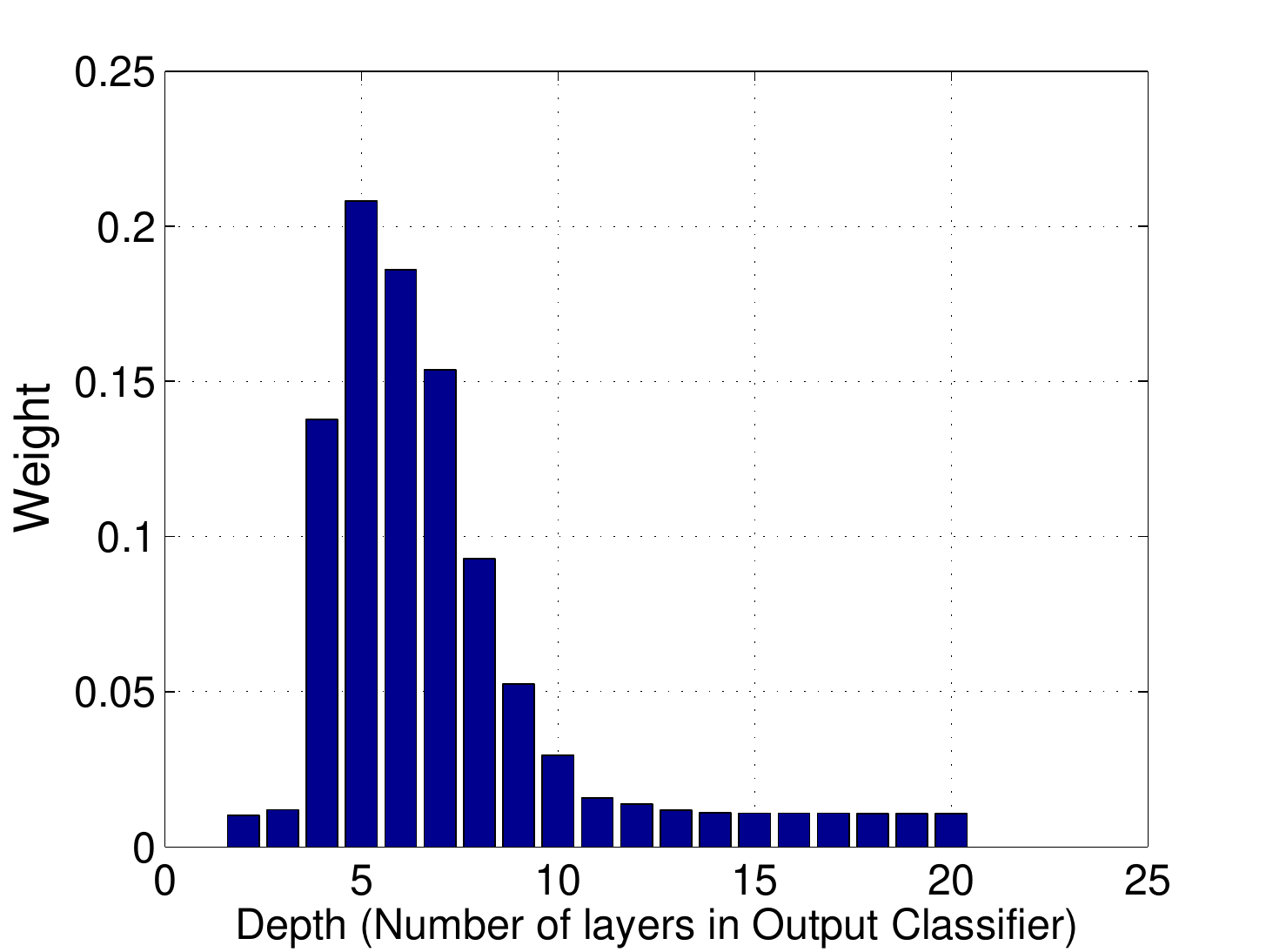}}
	\caption{Evolution of weight distribution of the classifiers over time using HBP on HIGGS dataset.}
	\label{fig:weightDistribution}
\end{figure*}

\subsection{Adapting the Effective Depth of the DNN}
In this section we look at the weight distribution learnt by HBP over time. We analyse the mean weight distribution in different segments of the Online Learning phase in Figure \ref{fig:weightDistribution}. In the initial phase (first 0.5\%), the maximum weight has gone to the shallowest classifier (with just one hidden layer). In the second phase (10-15\%), slightly deeper classifiers (classifiers with 4-5 layers) have picked up some weight, and in the third segment (60-80\%), even deeper classifiers have gotten more weight (classifiers with 5-7 layers). The shallow and the very deep classifiers receive little weight in the last segment showing HBPs ability to perform model selection. Few classifiers having similar depth indicates that the intermediate features learnt are themselves of discriminatory nature, which are being used by the deeper classifiers to potentially learn better features.

\subsection{Performance in Different Learning Stages}

\begin{figure}[H]
	\centering
	\subfigure[Error in 10-15\% of data]{
		\includegraphics[width=.18\textwidth]{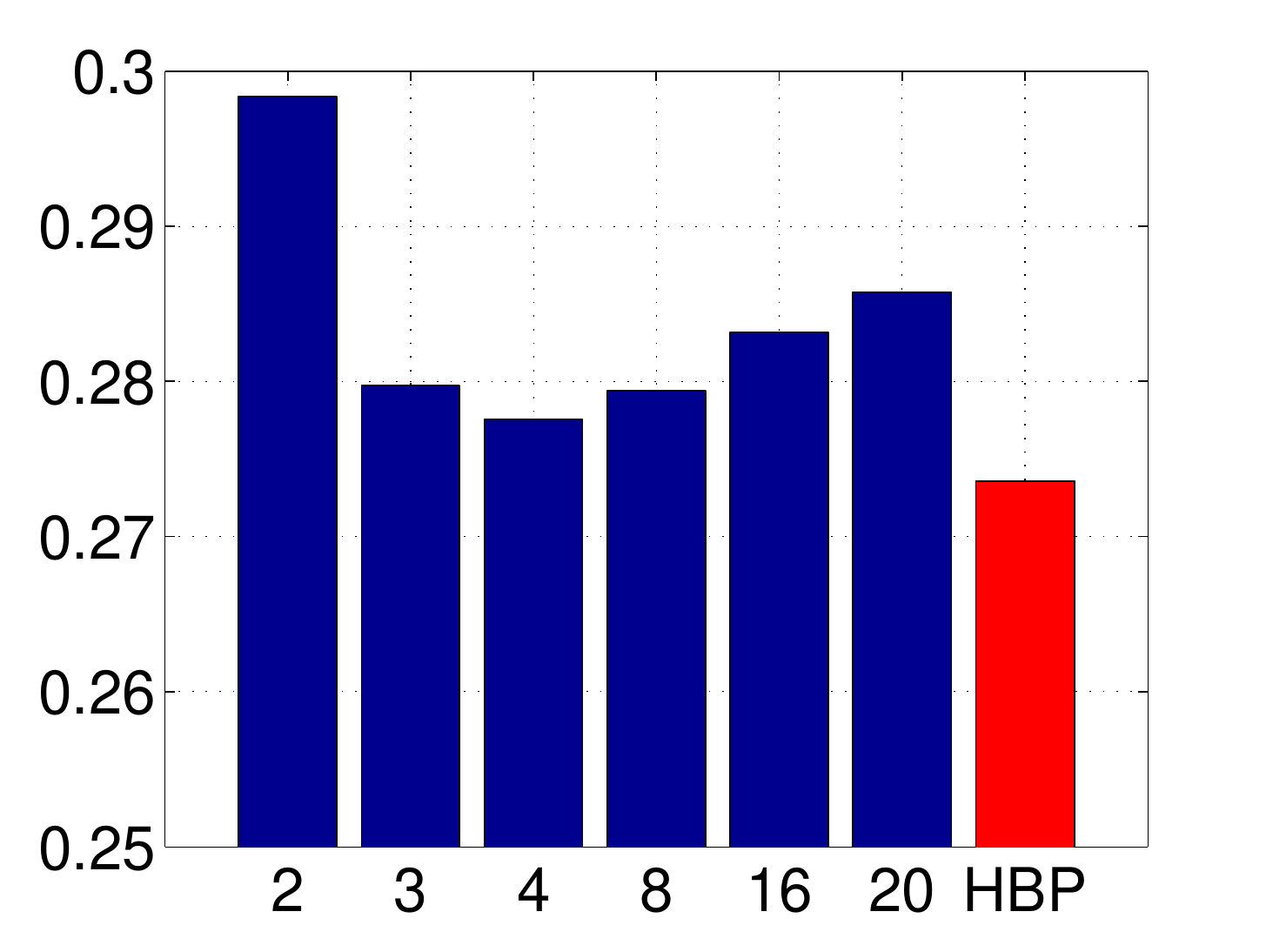}}
	\subfigure[Error in 60-80\% of data]{
		\includegraphics[width=.18\textwidth]{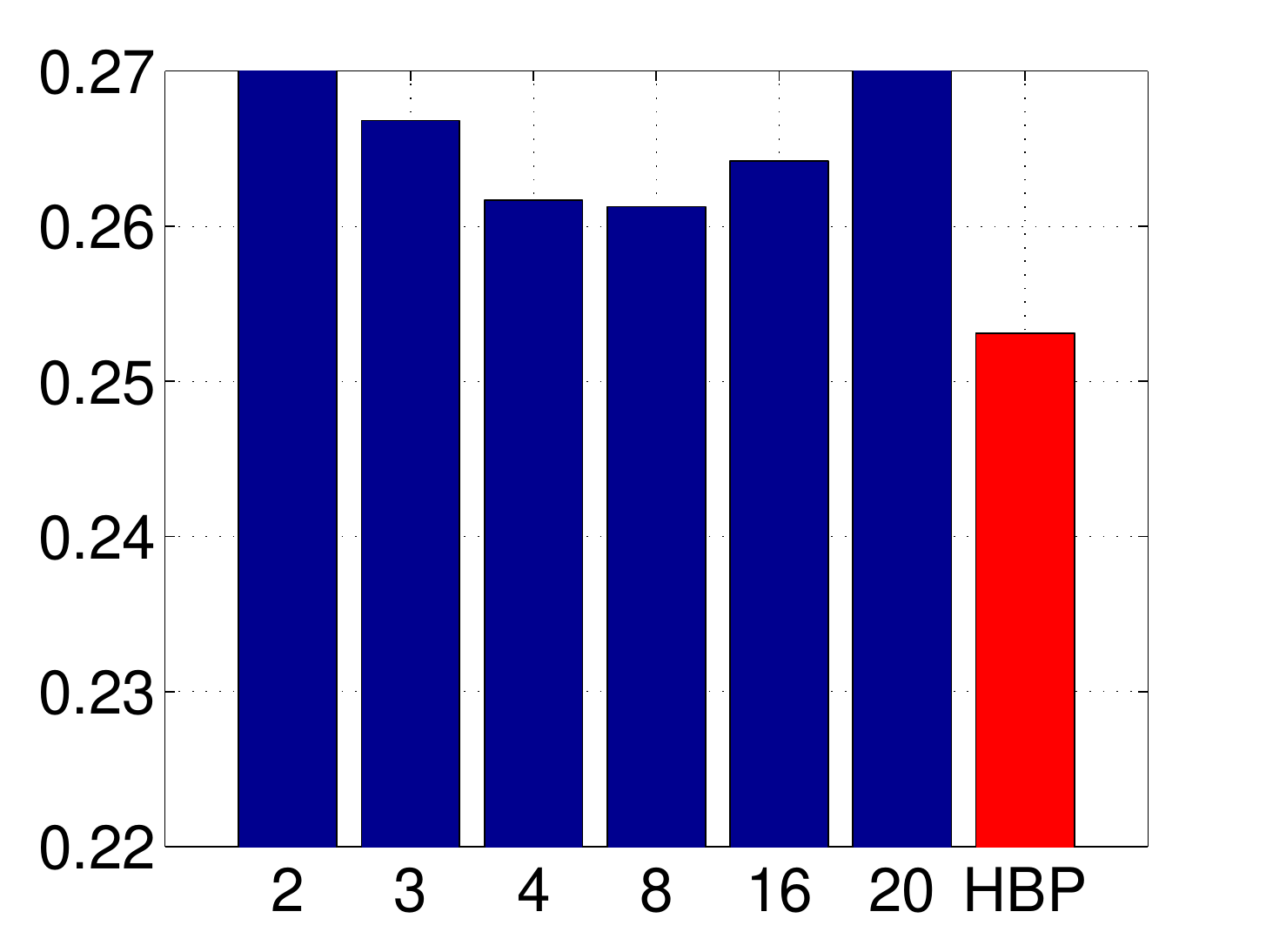}}
	\caption{Error Rate in different segments of the Data. Red represents HBP using a 20-layer network. Blue are OGD using DNN with layers = 2,3,4,8, 16 and 20.}
	\label{fig:window_higgs}
\end{figure}

We evaluate HBP performance in different segments of the data to see how the proposed HBP algorithm performed as compared to the DNNs of different depth in different stages of learning. In Figure \ref{fig:window_higgs}, we can see, HBP matches (and even beats) the performance of the best depth network in both the beginning and at a later stage of the training phase. This shows its ability to exploit faster convergence of shallow networks in the beginning, and power of deep representation towards the end. Not only is it able to do automatic model selection, but also it is able to offer a good initialization for the deeper representation, so that the depth of the network can be exploited sooner, thus beating a DNN of every depth.

\vspace{-0.5cm}
\subsection{Robustness to Depth of Base-Network}
We evaluate HBP performance with varying depth of the base network. We consider 12, 16, 20, and a 30-layer DNNs trained using HBP and compare their performance on Higgs against simple Online BP. See Table \ref{tab:depthRobustness} for the results, where the performance variation with depth does not significantly alter HBPs performance, while for simple Online BP, significant increase in depth hurts the learning process.

\begin{table}[H]
	\small
	\centering
	\caption{Robustness of HBP to depth of the base network compared to traditional DNN}
	\begin{tabular}{|c|cccc|}
		\hline
		\textbf{Depth} & \textbf{12} & \textbf{16} & \textbf{20} & \textbf{30} \\
		\hline
		{\textbf{Online BP}} & 0.2692 & 0.2731 & 0.2868 & 0.4770 \\
		\textbf{HBP} & 0.2609 & 0.2613 & 0.2615 & 0.2620 \\
		\hline
	\end{tabular}%
	\label{tab:depthRobustness}%
\end{table}%

\section{Conclusion}
We identified several issues which prevented existing DNNs from being used in an online setting, which meant that they could not be used for streaming data, and necessarily required storage of the entire data in memory. These issues arose from difficulty in model selection (appropriate depth DNN), and convergence difficulties from vanishing gradient and diminishing feature reuse. We used the "shallow to deep" principle, and designed Hedge Backpropagation, which enabled the usage of Deep Neural Networks in an online setting. HBP used a hedging strategy to make predictions with multiple outputs from different hidden layers of the network, and the backpropagation algorithm was modified to allow for knowledge sharing among the deeper and shallower networks. This approach automatically identified how and when to modify the effective network capacity in a data-drive manner, based on the observed data complexity. We validated the proposed method through extensive experiments on large datasets.

\bibliographystyle{aaai}
\bibliography{references}

\begin{thebibliography}{}

\bibitem[\protect\citeauthoryear{Alvarez and
  Salzmann}{2016}]{alvarez2016learning}
Alvarez, J.~M., and Salzmann, M.
\newblock 2016.
\newblock Learning the number of neurons in deep networks.
\newblock In {\em NIPS}.

\bibitem[\protect\citeauthoryear{Auer \bgroup et al\mbox.\egroup
  }{2002}]{auer2002nonstochastic}
Auer, P.; Cesa-Bianchi, N.; Freund, Y.; and Schapire, R.~E.
\newblock 2002.
\newblock The nonstochastic multiarmed bandit problem.
\newblock {\em SIAM Journal on Computing} 32(1):48--77.

\bibitem[\protect\citeauthoryear{Bengio, Courville, and
  Vincent}{2013}]{bengio2013representation}
Bengio, Y.; Courville, A.; and Vincent, P.
\newblock 2013.
\newblock Representation learning: A review and new perspectives.
\newblock {\em IEEE transactions on pattern analysis and machine intelligence}
  35(8):1798--1828.

\bibitem[\protect\citeauthoryear{Bengio, Goodfellow, and
  Courville}{2015}]{bengio2015deep}
Bengio, Y.; Goodfellow, I.~J.; and Courville, A.
\newblock 2015.
\newblock Deep learning.
\newblock {\em An MIT Press book in preparation. Draft chapters available at
  http://www. iro. umontreal. ca/~ bengioy/dlbook}.

\bibitem[\protect\citeauthoryear{Cesa-Bianchi and
  Lugosi}{2006}]{Cesa-Bianchi2006}
Cesa-Bianchi, N., and Lugosi, G.
\newblock 2006.
\newblock {\em Prediction, learning, and games}.
\newblock Cambridge University Press.

\bibitem[\protect\citeauthoryear{Chen, Goodfellow, and
  Shlens}{2016}]{chen2015net2net}
Chen, T.; Goodfellow, I.; and Shlens, J.
\newblock 2016.
\newblock Net2net: Accelerating learning via knowledge transfer.
\newblock {\em ICLR}.

\bibitem[\protect\citeauthoryear{Chollet}{2015}]{chollet2015keras}
Chollet, F.
\newblock 2015.
\newblock Keras.
\newblock \url{https://github.com/fchollet/keras}.

\bibitem[\protect\citeauthoryear{Choromanska \bgroup et al\mbox.\egroup
  }{2015}]{choromanska2015loss}
Choromanska, A.; Henaff, M.; Mathieu, M.; Arous, G.; and LeCun, Y.
\newblock 2015.
\newblock The loss surfaces of multilayer networks.
\newblock In {\em AISTATS}.

\bibitem[\protect\citeauthoryear{Crammer \bgroup et al\mbox.\egroup
  }{2006}]{Crammer2006}
Crammer, K.; Dekel, O.; Keshet, J.; Shalev-Shwartz, S.; and Singer, Y.
\newblock 2006.
\newblock Online passive-aggressive algorithms.
\newblock {\em JMLR}.

\bibitem[\protect\citeauthoryear{Dauphin \bgroup et al\mbox.\egroup
  }{2014}]{dauphin2014identifying}
Dauphin, Y.; Pascanu, R.; Gulcehre, C.; Cho, K.; Ganguli, S.; and Bengio, Y.
\newblock 2014.
\newblock Identifying and attacking the saddle point problem in
  high-dimensional non-convex optimization.
\newblock In {\em NIPS}.

\bibitem[\protect\citeauthoryear{Dredze, Crammer, and
  Pereira}{2008}]{dredze2008confidence}
Dredze, M.; Crammer, K.; and Pereira, F.
\newblock 2008.
\newblock Confidence-weighted linear classification.
\newblock In {\em ICML}.

\bibitem[\protect\citeauthoryear{Freund and Schapire}{1997}]{Freund1997}
Freund, Y., and Schapire, R.
\newblock 1997.
\newblock A decision-theoretic generalization of on-line learning and an
  application to boosting.
\newblock {\em Journal of computer and system sciences}.

\bibitem[\protect\citeauthoryear{Freund and
  Schapire}{1999}]{freund1999adaptive}
Freund, Y., and Schapire, R.
\newblock 1999.
\newblock Adaptive game playing using multiplicative weights.
\newblock {\em Games and Economic Behavior}.

\bibitem[\protect\citeauthoryear{Gama \bgroup et al\mbox.\egroup
  }{2014}]{gamajo2014survey}
Gama, J.; Zliobaite, I.; Bifet, A.; Pechenizkiy, M.; and Bouchachia, A.
\newblock 2014.
\newblock A survey on concept drift adaptation.
\newblock {\em ACM Computing Surveys (CSUR)} 46(4):44.

\bibitem[\protect\citeauthoryear{Gulcehre \bgroup et al\mbox.\egroup
  }{2016}]{gulcehre2016mollifying}
Gulcehre, C.; Moczulski, M.; Visin, F.; and Bengio, Y.
\newblock 2016.
\newblock Mollifying networks.
\newblock {\em preprint arXiv:1608.04980}.

\bibitem[\protect\citeauthoryear{He \bgroup et al\mbox.\egroup
  }{2016}]{he2015deep}
He, K.; Zhang, X.; Ren, S.; and Sun, J.
\newblock 2016.
\newblock Deep residual learning for image recognition.
\newblock {\em CVPR}.

\bibitem[\protect\citeauthoryear{Hoi \bgroup et al\mbox.\egroup
  }{2013}]{Hoi2013}
Hoi, S.; Jin, R.; Zhao, P.; and Yang, T.
\newblock 2013.
\newblock Online multiple kernel classification.
\newblock {\em Machine Learning}.

\bibitem[\protect\citeauthoryear{Hoi, Wang, and Zhao}{2014}]{hoi2014libol}
Hoi, S.; Wang, J.; and Zhao, P.
\newblock 2014.
\newblock Libol: A library for online learning algorithms.
\newblock {\em JMLR}.

\bibitem[\protect\citeauthoryear{Huang \bgroup et al\mbox.\egroup
  }{2016}]{huang2016deep}
Huang, G.; Sun, Y.; Liu, Z.; Sedra, D.; and Weinberger, K.
\newblock 2016.
\newblock Deep networks with stochastic depth.
\newblock {\em arXiv preprint arXiv:1603.09382}.

\bibitem[\protect\citeauthoryear{Ioffe and Szegedy}{2015}]{ioffe2015batch}
Ioffe, S., and Szegedy, C.
\newblock 2015.
\newblock Batch normalization: Accelerating deep network training by reducing
  internal covariate shift.
\newblock {\em arXiv preprint arXiv:1502.03167}.

\bibitem[\protect\citeauthoryear{Jin, Hoi, and Yang}{2010}]{Jin2010}
Jin, R.; Hoi, S.; and Yang, T.
\newblock 2010.
\newblock Online multiple kernel learning: Algorithms and mistake bounds.
\newblock In {\em Algorithmic Learning Theory}. Springer Berlin Heidelberg.

\bibitem[\protect\citeauthoryear{Kivinen, Smola, and
  Williamson}{2004}]{Kivinen2004}
Kivinen, J.; Smola, A.; and Williamson, R.
\newblock 2004.
\newblock Online learning with kernels.
\newblock {\em IEEE TSP} 52(8):2165--2176.

\bibitem[\protect\citeauthoryear{Krizhevsky, Sutskever, and
  Hinton}{2012}]{krizhevsky2012imagenet}
Krizhevsky, A.; Sutskever, I.; and Hinton, G.~E.
\newblock 2012.
\newblock Imagenet classification with deep convolutional neural networks.
\newblock In {\em Advances in neural information processing systems},
  1097--1105.

\bibitem[\protect\citeauthoryear{Larsson, Maire, and
  Shakhnarovich}{2016}]{larsson2016fractalnet}
Larsson, G.; Maire, M.; and Shakhnarovich, G.
\newblock 2016.
\newblock Fractalnet: Ultra-deep neural networks without residuals.
\newblock {\em arXiv preprint arXiv:1605.07648}.

\bibitem[\protect\citeauthoryear{LeCun, Bengio, and
  Hinton}{2015}]{lecun2015deep}
LeCun, Y.; Bengio, Y.; and Hinton, G.
\newblock 2015.
\newblock Deep learning.
\newblock {\em Nature} 521(7553):436--444.

\bibitem[\protect\citeauthoryear{Lee \bgroup et al\mbox.\egroup
  }{2015}]{lee2015deeply}
Lee, C.-Y.; Xie, S.; Gallagher, P.; Zhang, Z.; and Tu, Z.
\newblock 2015.
\newblock Deeply-supervised nets.
\newblock In {\em AISTATS}, volume~2, ~6.

\bibitem[\protect\citeauthoryear{Lee \bgroup et al\mbox.\egroup
  }{2016}]{lee2016dual}
Lee, S.-W.; Lee, C.-Y.; Kwak, D.-H.; Kim, J.; Kim, J.; and Zhang, B.-T.
\newblock 2016.
\newblock Dual-memory deep learning architectures for lifelong learning of
  everyday human behaviors.
\newblock In {\em IJCAI},  1669--1675.

\bibitem[\protect\citeauthoryear{Lee \bgroup et al\mbox.\egroup
  }{2017}]{lee2017lifelong}
Lee, J.; Yun, J.; Hwang, S.; and Yang, E.
\newblock 2017.
\newblock Lifelong learning with dynamically expandable networks.
\newblock {\em arXiv preprint arXiv:1708.01547}.

\bibitem[\protect\citeauthoryear{Loosli, Canu, and
  Bottou}{2007}]{loosli2007training}
Loosli, G.; Canu, S.; and Bottou, L.
\newblock 2007.
\newblock Training invariant support vector machines using selective sampling.
\newblock {\em Large scale kernel machines}  301--320.

\bibitem[\protect\citeauthoryear{Lu \bgroup et al\mbox.\egroup
  }{2015a}]{lu2015large}
Lu, J.; Hoi, S.; Wang, J.; Zhao, P.; and Liu, Z.
\newblock 2015a.
\newblock Large scale online kernel learning.
\newblock {\em J. Mach. Learn. Res}.

\bibitem[\protect\citeauthoryear{Lu \bgroup et al\mbox.\egroup
  }{2015b}]{lu2015budget}
Lu, J.; Hoi, S.~C.; Sahoo, D.; and Zhao, P.
\newblock 2015b.
\newblock Budget online multiple kernel learning.
\newblock {\em arXiv preprint arXiv:1511.04813}.

\bibitem[\protect\citeauthoryear{Nair and Hinton}{2010}]{nair2010rectified}
Nair, V., and Hinton, G.~E.
\newblock 2010.
\newblock Rectified linear units improve restricted boltzmann machines.
\newblock In {\em Proceedings of the 27th International Conference on Machine
  Learning (ICML-10)},  807--814.

\bibitem[\protect\citeauthoryear{Rosenblatt}{1958}]{Rosenblatt1958}
Rosenblatt, F.
\newblock 1958.
\newblock The perceptron: a probabilistic model for information storage and
  organization in the brain.
\newblock {\em Psychological review} 65(6):386.

\bibitem[\protect\citeauthoryear{Sahoo, Hoi, and Li}{2014}]{Sahoo2014}
Sahoo, D.; Hoi, S.; and Li, B.
\newblock 2014.
\newblock Online multiple kernel regression.
\newblock In {\em ACM SIGKDD}.

\bibitem[\protect\citeauthoryear{Sahoo, Hoi, and Zhao}{2016}]{sahoo2016cost}
Sahoo, D.; Hoi, S.; and Zhao, P.
\newblock 2016.
\newblock Cost sensitive online multiple kernel classification.
\newblock In {\em Asian Conference on Machine Learning},  65--80.

\bibitem[\protect\citeauthoryear{Shalev-Shwartz}{2007}]{Shalev-Shwartz2007}
Shalev-Shwartz, S.
\newblock 2007.
\newblock Online learning: Theory, algorithms, and applications.

\bibitem[\protect\citeauthoryear{Srinivas and
  Babu}{2015}]{srinivas2015learning}
Srinivas, S., and Babu, R.~V.
\newblock 2015.
\newblock Learning neural network architectures using backpropagation.
\newblock {\em arXiv preprint arXiv:1511.05497}.

\bibitem[\protect\citeauthoryear{Srivastava, Greff, and
  Schmidhuber}{2015}]{srivastava2015training}
Srivastava, R.~K.; Greff, K.; and Schmidhuber, J.
\newblock 2015.
\newblock Training very deep networks.
\newblock In {\em Advances in neural information processing systems},
  2377--2385.

\bibitem[\protect\citeauthoryear{Wei \bgroup et al\mbox.\egroup
  }{2016}]{wei2016network}
Wei, T.; Wang, C.; Rui, R.; and Chen, C.~W.
\newblock 2016.
\newblock Network morphism.
\newblock {\em ICML}.

\bibitem[\protect\citeauthoryear{Wu \bgroup et al\mbox.\egroup
  }{2017}]{wu2017sol}
Wu, Y.; Hoi, S.~C.; Liu, C.; Lu, J.; Sahoo, D.; and Yu, N.
\newblock 2017.
\newblock Sol: A library for scalable online learning algorithms.
\newblock {\em Neurocomputing}.

\bibitem[\protect\citeauthoryear{Zhou, Sohn, and Lee}{2012}]{zhou2012online}
Zhou, G.; Sohn, K.; and Lee, H.
\newblock 2012.
\newblock Online incremental feature learning with denoising autoencoders.
\newblock {\em AISTATS} 1001:48109.

\bibitem[\protect\citeauthoryear{Zinkevich}{2003}]{Zinkevich2003}
Zinkevich, M.
\newblock 2003.
\newblock Online convex programming and generalized infinitesimal gradient
  ascent.

\bibitem[\protect\citeauthoryear{Zoph and Le}{2016}]{zoph2016neural}
Zoph, B., and Le, Q.~V.
\newblock 2016.
\newblock Neural architecture search with reinforcement learning.
\newblock {\em arXiv preprint arXiv:1611.01578}.

\end{thebibliography}

\end{document}